\def\year{2021}\relax
\documentclass[letterpaper]{article} % DO NOT CHANGE THIS
\usepackage{aaai21}  % DO NOT CHANGE THIS
\usepackage{times}  % DO NOT CHANGE THIS
\usepackage{helvet} % DO NOT CHANGE THIS
\usepackage{courier}  % DO NOT CHANGE THIS
\usepackage[hyphens]{url}  % DO NOT CHANGE THIS
\usepackage{graphicx} % DO NOT CHANGE THIS
\usepackage{natbib}  % DO NOT CHANGE THIS AND DO NOT ADD ANY OPTIONS TO IT
\usepackage{caption} % DO NOT CHANGE THIS AND DO NOT ADD ANY OPTIONS TO IT
\usepackage{array}
\usepackage{cite}

\usepackage{mdwmath}

\usepackage{graphicx}         % standard LaTeX graphics tool
\usepackage{subfigure}
\usepackage{pifont}
\usepackage{bbm}
\usepackage{dsfont}

\usepackage{bm}
\usepackage{comment}
\usepackage{amsmath}
\usepackage{enumerate}  %\begin{enumerate}[i] % reacts to 1, a, A, i, and I
\usepackage{mathrsfs}
\usepackage{makeidx}         % allows index generation
\usepackage{graphicx}        % standard LaTeX graphics tool
\usepackage{subfigure}
\usepackage{epsfig,amssymb,latexsym}
\usepackage{psfrag}

\usepackage{algorithm}
\usepackage{algorithmicx}

\usepackage{subfigure}
\usepackage{algcompatible,lipsum}
\usepackage{cases}

\usepackage[misc]{ifsym}
\usepackage{fancyhdr}
\usepackage{multirow}

\usepackage{color}
\usepackage{indentfirst}

\usepackage{colortbl}
\definecolor{lightgray}{rgb}{.93,.93,.93}

\newcommand{\mr}{\mathrm}

\title{I3DOL: Incremental 3D Object Learning without Catastrophic Forgetting}
\author{
	Jiahua Dong\textsuperscript{\rm 1, \rm 2, \rm 3}, 
	Yang Cong\textsuperscript{\rm 1, \rm 2}\thanks{The corresponding author is Prof. Yang Cong.},
	Gan Sun\textsuperscript{\rm 1, \rm 2},
	Bingtao Ma\textsuperscript{\rm 1, \rm 2, \rm 3}
	and Lichen Wang\textsuperscript{\rm 4}\\
}
\affiliations{
    \textsuperscript{\rm 1}State Key Laboratory of Robotics, Shenyang Institute of Automation, Chinese Academy of Sciences, Shenyang, China.\thanks{This work is supported by the Major Project of the New Generation of Artificial Intelligence (2018AAA0102905) and National Nature Science Foundation of China under Grant (61722311, U1613214, 61821005, 62003336).}  \\
    
    \textsuperscript{\rm 2}Institutes for Robotics and Intelligent Manufacturing, Chinese Academy of Sciences, Shenyang, 110016, China.  \\
    
    \textsuperscript{\rm 3}University of Chinese Academy of Sciences, Beijing, 100049, China. \\
    
    \textsuperscript{\rm 4}Northeastern University, Boston, USA. \\

	\{dongjiahua1995, congyang81, sungan1412, mabingtao93, wanglichenxj\}@gmail.com
	
}
\begin{document}

\maketitle

\begin{abstract}
3D object classification has attracted appealing attentions in academic researches and industrial applications. However, most existing methods need to access the training data of past 3D object classes when facing the common real-world scenario: new classes of 3D objects arrive in a sequence. Moreover, the performance of advanced approaches degrades dramatically for past learned classes (\emph{i.e.}, catastrophic forgetting), due to the irregular and redundant geometric structures of 3D point cloud data. To address these challenges, we propose a new \underline{I}ncremental \underline{3D} \underline{O}bject \underline{L}earning (\emph{i.e.,} I3DOL) model, which is the first exploration to learn new classes of 3D object continually. Specifically, an adaptive-geometric centroid module is designed to construct discriminative local geometric structures, which can better characterize the irregular point cloud representation for 3D object. Afterwards, to prevent the catastrophic forgetting brought by redundant geometric information, a geometric-aware attention mechanism is developed to quantify the contributions of local geometric structures, and explore unique 3D geometric characteristics with high contributions for classes incremental learning. Meanwhile, a score fairness compensation strategy is proposed to further alleviate the catastrophic forgetting caused by unbalanced data between past and new classes of 3D object, by compensating biased prediction for new classes in the validation phase. Experiments on 3D representative datasets validate the superiority of our I3DOL framework.
\end{abstract}

\section{Introduction}
Object recognition technology has achieved remarkable developments in great quantifies of research fields, \emph{e.g.}, autonomous driving \cite{Behl_2017_ICCV, Li_Peixuan_RTM3D}, intelligent robotics \cite{8593741, ijcai2020-77, chen-etal-2020-bridging}, object clustering \cite{yang2020adversarial, ZhangCSWD20, ZhangGPVTFC}, medical diagnosis \cite{Semantic_Transferable_Dong_ICCV2019, What_Transferred_Dong_CVPR2020}, transfer learning \cite{CSCL_Dong_ECCV2020, DBLP:journals/corr/abs-1907-08375, DBLP:conf/ijcai/ZhangLFY0020, DBLP:journals/corr/abs-2008-01454, DBLP:journals/corr/abs-2006-13022} and federated learning \cite{wang2019eavesdrop, wang2020towards}. When compared with 2D vision, the unordered 3D point cloud representation collected by depth cameras or LiDAR system is more difficult for object recognition to characterize the 3D geometric information. To this end, various deep neural networks capable of reasoning about 3D geometric layout and structure \cite{Qi_2017_CVPR, NIPS2017_7095} are proposed to explore the task-specific semantic content. Intuitively, the advent of convolutional architectures \cite{NIPS2018_7362, Li_2018_CVPR, dgcnn} has dramatically boosted the performance of 3D object classification for point cloud representation.

\begin{figure}[t]
	\centering
	\includegraphics[width =235pt, height =168pt]{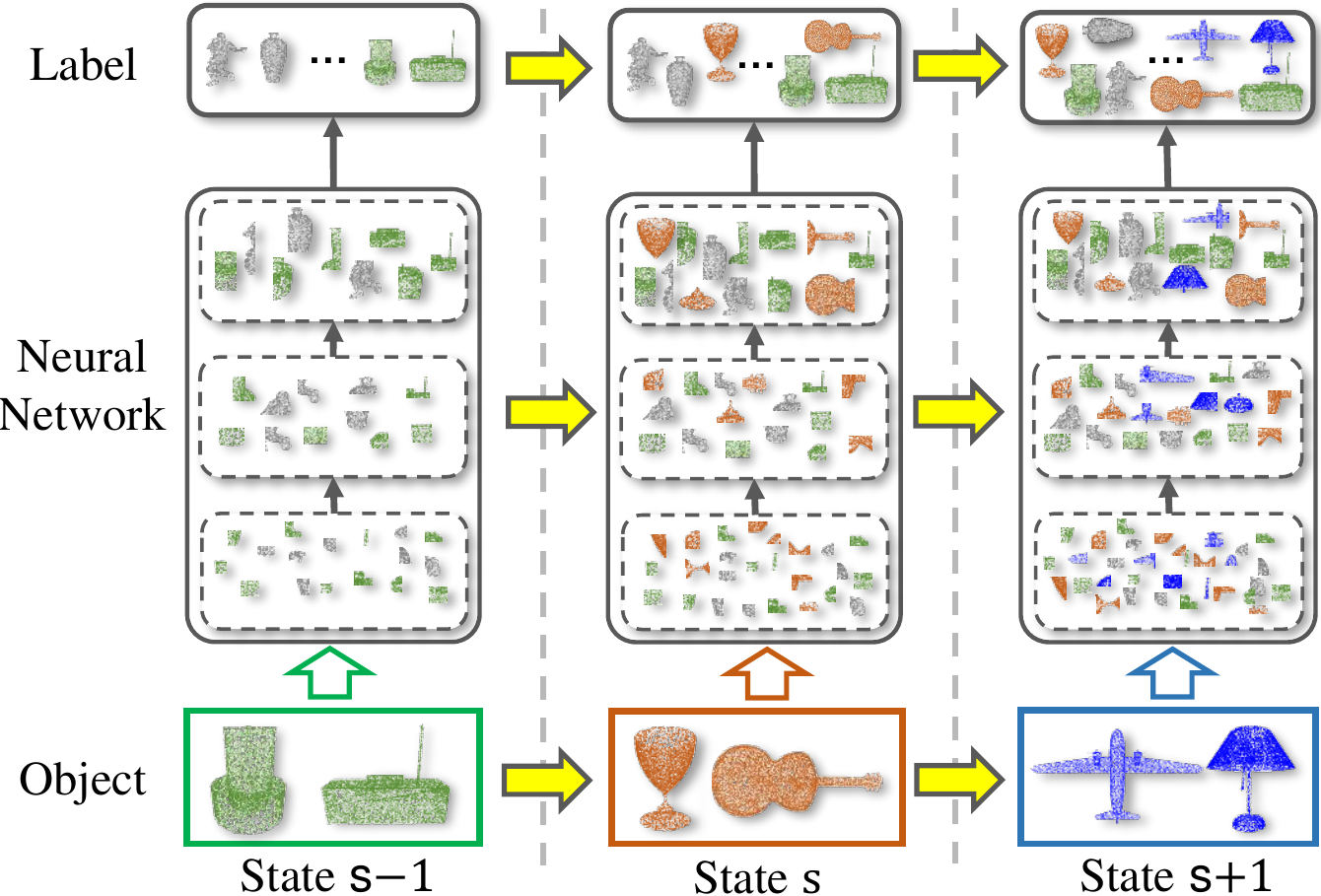}
	\caption{Demonstration of our I3DOL model to learn new classes of 3D objects consecutively.}
	\label{fig:motivation_of_our_model}
\end{figure}

However, these existing methods are trained on a prepared well-labeled dataset, whose the number of 3D object categories is fixed in advance. This setup significantly limits their application promotion in the real-world scenarios where new classes of 3D object arrive continually in a streaming manner, as shown in Figure~\ref{fig:motivation_of_our_model}. For example, housekeeping robots \cite{she2019openlorisobject} working for indoor tasks cannot perform well in outdoor scenes, due to the lack of continual learning capacity for new 3D objects. A trivial solution to address this is to access all the training data of past learned indoor object classes, when long delay is allowed for updating the current model. Nevertheless, it can inevitably result in high computational power and storage (\emph{e.g.}, large infrastructures), which are not satisfied in real-world scenarios. Besides, the straightforward way is to apply current classes incremental models \cite{Rebuffi_2017_CVPR, Castro_2018_ECCV, Wu_2019_CVPR, Oleksiy_2019_CVPR} in 2D vision equipped with a 3D point cloud feature extractor \cite{NIPS2017_7095, Qi_2017_CVPR} into learning new classes of 3D object. However, these existing methods cannot explore unique and informative 3D geometric characteristics for classes incremental learning and further cause catastrophic forgetting, due to the irregular and redundant geometric structures within 3D point cloud \cite{NIPS2017_7095} (\emph{e.g.}, tables with missed legs or deformable permutations). Therefore, learning new classes incrementally for 3D objects without retaining the training data of past classes is a crucial real-world challenge.

To tackle this challenge, we develop a new \underline{I}ncremental \underline{3D} \underline{O}bject \underline{L}earning (\emph{i.e.,} I3DOL) model, which intends to alleviate catastrophic forgetting for point cloud representation of past classes when learning new classes continually, as shown in Figure~\ref{fig:overview_of_our_model}. Specifically, by constructing discriminative local geometric structures of point cloud, an adaptive-geometric centroid module associated with an adaptive receptive field is developed to characterize the irregular point cloud representation. Meanwhile, a geometric-aware attention mechanism is designed to capture the intrinsic relationships between local geometric structures by quantifying the contributions of each local geometric structures for class incremental learning. In other words, it pays more attention on unique and informative local geometric structures to prevent the catastrophic forgetting while neglecting redundant geometric information in point cloud representation. To further alleviate the catastrophic forgetting for past learned 3D object classes, a score fairness compensation strategy is proposed to address the unbalanced training data between past learned and new classes of 3D object, which corrects the biased prediction for past classes via adaptive compensation in the validation phase. Experiments on several 3D benchmark classification tasks strongly demonstrate the effectiveness of our I3DOL model. In summary, the main contributions of this work are presented as follows: 
\begin{itemize}
	\item A new Incremental 3D Object Learning (I3DOL) model for point cloud representation is designed to learn new classes of 3D object continually while alleviating the catastrophic forgetting for past classes. To our best knowledge, this is the first exploration about classes incremental learning in the 3D object recognition field.
	
	\item We develop an adaptive-geometric centroid module to better characterize the irregular point cloud representation, which could construct several discriminative local geometric structures with an adaptive receptive field for each point cloud. 
	
	\item To prevent the catastrophic forgetting, a geometric-aware attention mechanism is designed to highlight beneficial 3D geometric characteristics with high contributions for classes incremental learning, while a score fairness compensation strategy is developed to compensate the biased score prediction in the validation stage. 
	
\end{itemize}

\section{Related Work}
This section reviews some related researches about 3D object classification and classes incremental learning.

\subsection{3D Object Classification}
Although enormous shape descriptors \cite{6619148, 7300438, 8237705, 4650967, doi:10.1111/j.1467-8659.2009.01515.x} are handcrafted by domain experts for point cloud representation, they are invariant to the specific transformations, and cannot generalize well to 3D objects recognition with various categories. After Qi \emph{et al.} propose a pioneer work PointNet \cite{Qi_2017_CVPR} to directly process point cloud data, the advent of deep network architecture has achieved impressive successes in 3D object classification due to its powerful characterization capacity. PointNet++ \cite{NIPS2017_7095} encodes multi-scale hierarchical semantic context of point cloud. Moreover, some permutation invariant architectures \cite{Li_2018_CVPR, NIPS2018_7362} are developed for deep learning with orderless point clouds, which explore the spatially-local correlations about data distribution. \cite{dgcnn} focus on capturing  topology information to enrich the characterization power of point cloud by incorporating with the dynamical graph. However, these approaches cannot be directly applied into practical applications where new classes of 3D object arrive continuously in a streaming manner.

\begin{figure*}[t]
	\centering
	\includegraphics[width =505pt, height =150pt]{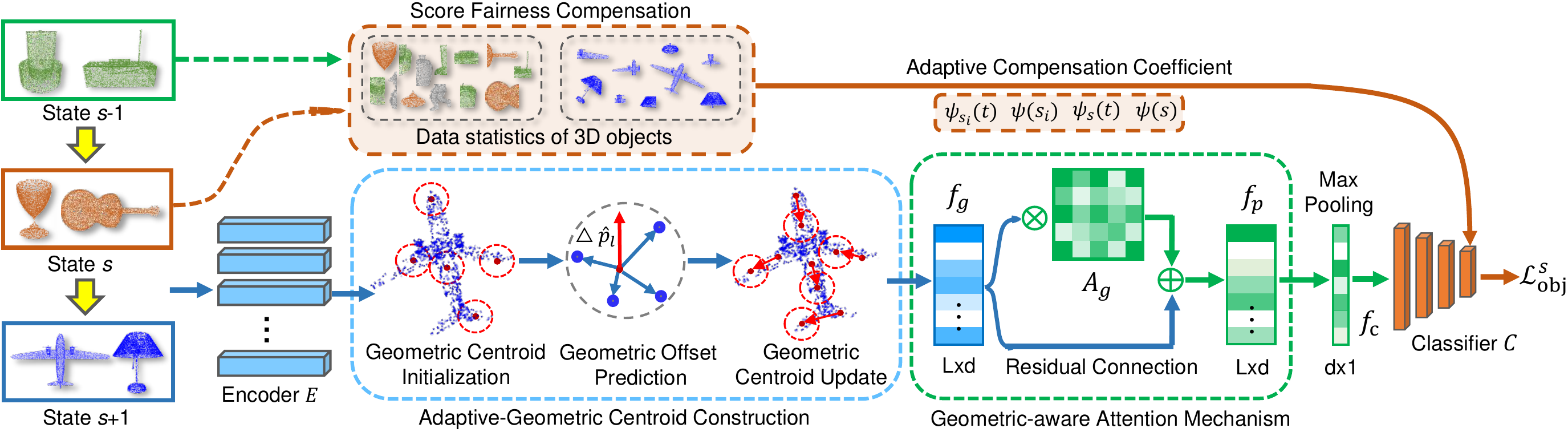}
	\caption{Overview of our I3DOL model, which mainly consists of the \textit{adaptive-geometric centroid construction} to explore several local geometric structures, the \textit{geometric-aware attention mechanism} to capture informative 3D geometric characteristics in local geometric structures with high contributions for classes incremental learning, and the \textit{score fairness compensation} to further alleviate the catastrophic forgetting brought by unbalanced training samples.} 
	\label{fig:overview_of_our_model}
\end{figure*}

\subsection{Classes Incremental Learning}
Generally, depending on whether nothing or synthetic data or real data from the past classes is available, the existing methods about classes incremental learning \cite{10.1007/978-3-319-46493-0_37, Kirkpatrick3521, NIPS2018_7836, Oleksiy_2019_CVPR,Wu_2019_CVPR, NIPS2019_9429} in 2D vision contain threefold division. Specifically, \cite{10.1007/978-3-319-46493-0_37, Shmelkov_2017_ICCV} employ knowledge distillation to prevent catastrophic forgetting for past classes without accessing their training data. \cite{Kirkpatrick3521} constrain the architecture weights of new tasks to maintain the better performance on past tasks. Furthermore, \cite{Oleksiy_2019_CVPR, 10.5555/3294996.3295059, DBLP:journals/corr/SeffBSL17, NIPS2018_7836} highly depend on the capability of generative adversarial networks to replay synthetic data for past classes. When a small number of exemplars from each old class are selected for training, \cite{Rebuffi_2017_CVPR, Castro_2018_ECCV, deesil-eccv2018} focus on alleviating the effect of unbalanced training samples between the past and new categories. \cite{Belouadah_2019_ICCV} design an additional memory with negligible added cost to record past classes statistics. \cite{Xiao2014ErrorDrivenIL, DBLP:journals/corr/RusuRDSKKPH16} propose to expand the network progressively as new training data arrives. \cite{Wu_2019_CVPR} correct the bias towards new classes brought by fully-connected layer for large-scale incremental learning. \cite{NIPS2019_9429} develop a random path selection to choose optimal paths for new tasks. However, these methods in 2D vision cannot be successfully applied into 3D object recognition, since the irregular and redundant geometric structures of point cloud representation make them difficult to explore unique 3D geometric characteristics that are beneficial for classes incremental learning.

\section{Our Proposed I3DOL Model}
\subsection{Problem Definition and Overview}
For classes incremental learning of 3D object, we follow the general experimental configuration in 2D vision \cite{Rebuffi_2017_CVPR, Castro_2018_ECCV, Wu_2019_CVPR, NIPS2019_9429}. There are total $S$ incremental states and the training data $D$ is denoted as $D = \{D_s\}_{s=1}^S$, where $D_s = \{x_i^s, y_i^s\}_{i=1}^{n_s}$ represents  $n_s$ point cloud data in the $s$-th incremental state. $x_i^s\in\mathbb{R}^{U\times 3}$ and $y_i^s$ denote the $i$-th point cloud data with 3 dimensional coordinates and its corresponding label, and $U$ is the number of sampling points for each 3D object. In the $s$-th incremental state, the labels in $D_s$ consist of $c_s$ new classes, which are different from $c_p = \sum_{i=1}^{s-1}c_i$ past classes in the previous $s-1$ incremental states. Similar to \cite{Rebuffi_2017_CVPR, Wu_2019_CVPR, NIPS2019_9429}, our goal is to make predictions for both $c_s$ new classes and $c_p$ past classes in the $s$-th incremental state, when the new coming data $D_s$ and the selected exemplars set $M$ from past classes are available. Note that $|M|$ is a small value when compared with $n_s$, and it satisfies $|M|/c_p\ll n_s/c_s$ in our experiments.

The overview framework of our I3DOL model is depicted in Figure~\ref{fig:overview_of_our_model}. Specifically, the point cloud data of new classes in the $s$-th incremental state is first forwarded into encoder $E$ to extract mid-level features. Then the adaptive-geometric centroid module constructs several discriminative local geometric structures to better characterize the irregular point cloud representation. Meanwhile, the geometric-aware attention module explores the unique and informative 3D geometric characteristics in local geometric structures with high contritions to alleviate catastrophic forgetting while neglecting the redundant information. Furthermore, we develop the score fairness compensation strategy to correct biased score prediction for new classes, which further prevent catastrophic forgetting caused by unbalanced data between past and new classes of 3D object.

\subsection{Adaptive-Geometric Centroid Construction}\label{sec:centroid_construction}
Suppose that each point cloud is composed of $L$ local geometric structures, which are regarded as $L$ point sets $\big\{G_l|G_l = \{\hat{p}_l, p_{l1}, \cdots, p_{lk}\in\mathbb{R}^3\}\big\}_{l=1}^L$. The $l$-th geometric region $G_l$ consists of a structural centroid $\hat{p}_l$ and its corresponding $k$ nearest neighbor points $\{p_{l1}, \cdots, p_{lk}\}$ surrounded around $\hat{p}_l$. Note that the location of $\hat{p}_l$ determines where the local geometric structure is and what the $k$ nearest neighbor points are included.
To extract local discriminative features, most previous methods \cite{NIPS2017_7095, NIPS2018_7362} directly utilize the farthest point sampling or random sampling to obtain the structural centroids for local geometric structures. Although these strategies can fully cover over the whole point cloud, the selected centroids cannot cover the structures with unique 3D geometric characteristics for both past and new classes, and neglect the common useless characteristics in classes incremental learning. Intuitively, the local geometric structures sharing common characteristics could result in the catastrophic forgetting for past classes, while the unique object components and geometric layouts promote to overcome it effectively.

Consequently, as depicted in Figure~\ref{fig:overview_of_our_model}, the adaptive-geometric centroid module with adaptive receptive field is developed to construct local geometric structures, which adjusts the selected structural centroids adaptively via geometric offset prediction along the training process. Different from deformable convolution \cite{DBLP:journals/corr/DaiQXLZHW17} which utilizes semantic features for offset prediction in 2D images, we consider the local edge vector of each geometric structure as a guidance for training. Specifically, the semantic knowledge of each edge is first transformed into the weight of edge vector, and then the weighted edge vectors are aggregated together to predict offset direction of the structural centroid. Generally, the learned offset is adaptively determined by the voting of surrounding edge vectors with different significances. After initializing the locations of $L$ structural centroids via the farthest point sampling \cite{NIPS2017_7095, NIPS2018_7362} over the point cloud, we collect the corresponding $k$ nearest neighbor points around each structural centroid to construct $L$ geometric structures. The offset prediction $\triangle\hat{p}_l$ for the $l$-th centroid $\hat{p}_l$ is:
\begin{equation}
\triangle \hat{p}_l = \frac{1}{k} \sum\limits_{i=1}^k \big(T_p((\hat{f}_l-f_{li}); \theta_{T_p}) \cdot (\hat{p}_l-p_{li}) \big),
\label{eq:offset_prediction}
\end{equation}
where $T_p$ denotes a convolutional layer with parameter as $\theta_{T_p}$, which adaptively transforms the semantic information into the weight of edge vector. $\hat{p}_l$ and $\{p_{li}\}_{i=1}^k$ denote the location of the $l$-th centroid and its $k$ nearest neighbor points, respectively. $\hat{f}_l$ and $\{f_{li}\}_{i=1}^k$ are the semantic features of $\hat{p}_l$ and $\{p_{li}\}_{i=1}^k$, which are extracted by the encoder $E$ in Figure~\ref{fig:overview_of_our_model}. $(\hat{p}_l-p_{li})$ is the direction of local edge vector with respect to the $l$-th centroid $\hat{p}_l$. With the offset prediction $\triangle\hat{p}_l$ in Eq.~\eqref{eq:offset_prediction}, we can update the $l$-th structural centroid $\hat{p}_l$ by adding the offset $\triangle\hat{p}_l$ back to $\hat{p}_l$, and reconstruct the $l$-th local geometric structure by searching $k$ new nearest neighbor points $\{p_{l1}, \cdots, p_{lk}\}$ around the new updated $\hat{p}_l$, \emph{i.e.},
\begin{equation}
\begin{split}
&\qquad\qquad\qquad\hat{p}_l = \hat{p}_l + \triangle\hat{p}_l, \\
&\{p_{l1}, \cdots, p_{lk}\} = \mr{kNN}(\hat{p}_l|p_j\in\mathbb{R}^3, j=1,\cdots, U),
\end{split}
\label{eq:local_structure_update}
\end{equation}
where $\mr{kNN}(\cdot)$ collects $k$ nearest neighbor points around the new updated centroid $\hat{p}_l$ by searching all points $\{p_j\}_{j=1}^U$ over the whole point cloud. Therefore, the semantic feature $\hat{f}_l$ of the $l$-th updated centroid $\hat{p}_l$ can be computed by gathering all points features inside the $l$-th updated local geometric structures, \emph{i.e.},
\begin{equation}
\hat{f}_l = \max_{i=1,2,\cdots,k} T_g(f_{li}; \theta_{T_g}),
\label{eq:update_centroid_feature}
\end{equation}
where $\{f_{li}\}_{i=1}^k$ represent the features of the updated $k$ nearest neighbor points. $T_g$ denotes a convolutional block to gather all points features, and the network weight is $\theta_{T_g}$. Similar to the concatenation strategy in \cite{NIPS2017_7095}, we concatenate all centroids features $\{\hat{f}_l\}_{l=1}^L$ from $L$ local geometric structures as the ultimate extracted features $f_g\in\mathbb{R}^{L\times d}$ over the whole point cloud, where $d$ is the feature dimension of each local geometric structure. $f_g$ is then forwarded into the geometric-aware attention module, as presented in Figure~\ref{fig:overview_of_our_model}.

\subsection{Geometric-Aware Attention Mechanism}\label{sec:geometric_aware_attention}
Although our adaptive-geometric centroid construction module could explore $L$ accurate structural centroids $\{\hat{p}_l\}_{l=1}^L$ with discriminative features $\{\hat{f}_l\}_{l=1}^L$, each centroid cannot contribute equally to explore informative 3D geometric characteristics for both past and new classes. In other words, some local geometric structures covering common characteristics can promote catastrophic forgetting for past learned classes, and the others with unique 3D characteristics prevent it efficiently. To this end, as shown in Figure~\ref{fig:overview_of_our_model}, we design the geometric-aware attention module to highlight unique 3D geometric characteristics that are beneficial for incremental learning, while preventing catastrophic forgetting caused by common characteristics. To be specific, it quantifies the contributions of different local geometric structures $\{G_l\}_{l=1}^L$ for classes incremental learning of 3D objects, and highlights those unique local geometric structures with high contributions to alleviate the catastrophic forgetting of past learned classes.

Motivated by the channel attention on feature responses \cite{zhang2018rcan} in 2D vision, we integrate a residual mechanism into the geometric-aware attention module to quantify the significance of local geometric structure. Then the ultimate semantic feature $f_p\in\mathbb{R}^{L\times d}$ over whole point cloud is:
\begin{equation}
\begin{split}
f_p &= A_g\cdot f_g + f_g \\
&= \Phi_1\big(T_u(\Phi_2(T_d(f_g; \theta_{T_d})); \theta_{T_u})\big)\cdot f_g + f_g,
\end{split}
\label{eq:centroid_attention}
\end{equation}
where $A_g = \Phi_1\big(T_u(\Phi_2(T_d(f_g; \theta_{T_d})); \theta_{T_u})\big)\in\mathbb{R}^{L\times d}$ denotes the learned geometric-aware attention, \emph{i.e.}, quantified contribution of each local geometric structure. $\Phi_1$ and $\Phi_2$ represent the sigmoid and ReLU activation functions, respectively. $T_d$ is a channel-downscaling convolutional block with the reduction rate as $r$, and $T_u$ is channel-upscaling convolutional layer with the increase ratio as $r$. $\theta_{T_u}$ and $\theta_{T_d}$ are the network parameters of $T_u$ and $T_d$. Note that we then utilize max pooling operation on $f_p$ to obtain the global feature $f_c\in\mathbb{R}^{d}$, before forwarding it into the classifier $C$ for performance prediction, as shown in Figure~\ref{fig:overview_of_our_model}.

\subsection{Score Fairness Compensation}\label{sec:dual_adaptive_fairness_compensations}
Even though the discriminative features of local geometric structures are extracted via above subsections, the classifier $C$ is prone to the catastrophic forgetting due to the unbalanced data distributions between the past and new classes of 3D objects. Most existing models \cite{Castro_2018_ECCV, Rebuffi_2017_CVPR, Wu_2019_CVPR, NIPS2019_9429} in 2D vision utilize the knowledge distillation to address this challenge. However, they cannot alleviate the problem that classifier $C$ tends to predict past 3D objects as new classes. Particularly, the classifier $C$ has higher preference for great quantities of new 3D objects classes. Obviously, the important factor causing catastrophic forgetting for past 3D objects is the highly biased probability prediction in the last layer of classifier $C$. To tackle this issue, the score fairness compensation strategy is developed to maintain fairness between past and new classes of 3D objects in classifier $C$ by compensating biased prediction for new 3D objects in the validation phase.

\begin{table*}[t]
	\centering
	\setlength{\tabcolsep}{2.61mm}
	\caption{Quantitative comparisons on ModelNet dataset \cite{7298801} with an increment of 4 classes.}
	\scalebox{0.94}{
		\begin{tabular}{|c|cccccccccc|c|}
			\hline
			Comparison Methods & 4 & 8 & 12 & 16 & 20 & 24 & 28 & 32 & 36 & 40 & Avg \\
			\hline
			LwF \cite{10.1007/978-3-319-46493-0_37}	& 96.5 & 87.2 & 77.5 & 70.6 & 62.3 & 56.8 & 44.7 & 39.4 & 36.1 & 31.5 & 60.3 \\
			iCaRL \cite{Rebuffi_2017_CVPR} & 96.8 & 90.4 & 83.6 & 78.3 & 72.5 & 67.3 & 59.6 & 53.1 & 47.8 & 39.6 & 68.9 \\
			DeeSIL \cite{deesil-eccv2018} & 97.7 & 91.5 & 85.4 & 80.5 & 74.4 & 71.8 & 65.3 & 58.7 & 52.4 & 43.7 & 72.1 \\
			EEIL \cite{Castro_2018_ECCV} & 97.6 & 93.8 & 87.5 & 81.6 & 78.2 &  74.7 & 69.2 & 62.4 & 56.8 & 48.1 & 75.0 \\
			IL2M \cite{Belouadah_2019_ICCV} & 97.8 & 95.1 & 89.4 & 85.7 & 83.8 & 82.2 & 78.4 & 72.8 & 67.9 & 57.6 & 81.1 \\
			DGMw \cite{Oleksiy_2019_CVPR} & 97.5 & 93.2 & 86.4 & 82.5 & 80.1 & 78.4 & 73.6 & 65.3 & 61.5 & 53.4 & 77.2  \\
			DGMa \cite{Oleksiy_2019_CVPR} & 97.5 & 93.4 & 84.7 & 81.8 & 79.5 & 77.8 & 74.1 & 67.4 & 60.8 & 51.5 & 76.8 \\
			BiC \cite{Wu_2019_CVPR} & 97.8 & 95.5 & 88.5 & 86.9 & 84.3 & 83.1 & 79.3 & 74.2 & 70.7 & 59.2 & 82.0 \\
			RPS-Net \cite{NIPS2019_9429} & 97.7 & 94.6 & 90.3 & 88.2 & 86.7 & 82.5 & 78.0 & 73.6 & 68.4 & 58.3 & 81.7 \\
			\hline
			\hline
			\rowcolor{lightgray}
			Ours-w/oAG & 97.8 & 94.3 & 90.1 & 87.5 & 84.2 & 81.7 & 77.9 & 73.5 & 68.4 & 59.1 & 81.5  \\
			
			\rowcolor{lightgray}
			Ours-w/oGA & 98.1 & 96.0 & 92.4 & 89.7 & 88.2 & 84.5 & 81.3 & 74.0 & 71.7 & 60.3 & 83.6  \\
			
			\rowcolor{lightgray}
			Ours-w/oSF & \textbf{98.2} & 96.1 & 92.0 & 90.3 & 88.9 & 85.4 & 80.7 & 75.8 & 72.4 & 60.8 & 84.1 \\
			
			\rowcolor{lightgray}
			Ours & 98.1 & \textbf{97.0} & \textbf{93.4} & \textbf{91.1} & \textbf{89.7} & \textbf{88.2} & \textbf{83.5} & \textbf{77.8} & \textbf{73.1} & \textbf{61.5} & \textbf{85.3} \\
			\hline					
		\end{tabular}
	} 	
	\label{tab:exp_modelnet40_dataset}
\end{table*}

The prediction bias for new classes of 3D objects appears during the inference stage, due to the unbalanced training samples. To this end, in the validation phase, we correct the prediction probability bias of past classes by incorporating with the initial data statistics of past classes obtained when they were initially learned in the training stage. The intuitive explanation is that prediction model is more confident when all training data of past 3D objects is available. Moreover, the initial data statistics are available across all incremental states, and the memory storage for them can be negligibly small. Concretely, for the $t$-th past learned class, the predicted probability via classifier $C$ is rectified by:
\begin{equation}
\begin{split}
C^s(f_c; \theta_{C})^t =
\left\{
\begin{aligned}		
&C(f_c; \theta_{C})^t \cdot \frac{\psi_{s_i}(t)}{\psi_{s}(t)} \cdot \frac{\psi(s)}{\psi(s_i)}, \text{if new classes},  \\
&C(f_c; \theta_{C})^t, \text{otherwise}, \\
\end{aligned} 					
\right.											
\end{split}							
\label{eq:score_fairness_compensation}	
\end{equation}
where $C^s(f_c; \theta_{C})^t$ and $C(f_c; \theta_{C})^t$ represent the probabilities of the $t$-th classes with the fairness compensation in validation stage or not. $\psi_{s_i}(t)$ and $\psi_s(t)$ denote the average classification scores predicted as the $t$-th class in the initial state $s_i$ and current incremental state $s$. Note that all training data of the $t$-th past 3D object class first appears in the initial state $s_i$. $\psi(s_i)$ and $\psi(s)$ are the mean prediction scores for all new coming classes of 3D objects in the states $s_i$ and $s$. Moreover, Eq.~\eqref{eq:score_fairness_compensation} applies the probability rectification into the past classes predictions only when the point cloud is initially predicted as the new classes. Obviously, by rescaling the predicted probabilities of past 3D objects classes with an adaptive statistic coefficient $\frac{\psi_{s_i}(t)}{\psi_{s}(t)} \cdot \frac{\psi(s)}{\psi(s_i)}$, Eq.~\eqref{eq:score_fairness_compensation} facilitates the inference fairness between the past and new classes of 3D objects.

\begin{algorithm}[t]
	\caption{\small Optimization Framework of Our I3DOL Model.}
	\begin{algorithmic}[1]
		\State {\bfseries Input:} The training data $D = \{D_s\}_{s=1}^S$ and the number of examples $|M|$. 
		\State {\bfseries Initialize: $\{\theta_{D_s}\}_{s=1}^S$};
		\State {\bfseries For} $s=1, \cdots, S$ \textbf{do}
		\State \quad Update exemplars set $M$;
		\State {\quad\bfseries While} not converged \textbf{do}
		\State \qquad Randomly sample a batch of examples from the new coming data $D_s$ and the exemplars set $M$;
		\State \qquad Update $\theta_{D_s}$ via Eq.~\eqref{eq:overall_training_objective};
		\State {\quad\bfseries End}
		\State \quad Store data statistics for score fairness compensation of Eq.~\eqref{eq:score_fairness_compensation} in the validation phase;
		\State {\bfseries End}
		\State {\bfseries Return} $\{\theta_{D_s}\}_{s=1}^S$;
	\end{algorithmic}
	\label{alg:I3DOL}
\end{algorithm}

\subsection{Implementation Details}
For the configuration of network architecture, we employ PointNet \cite{Qi_2017_CVPR} as the backbone framework of encoder $E$, and apply four-layer multi-layer perception as classifier $C$. Furthermore, we utilize the Adam optimizer for model optimization, where the learning rate and weight decay are initialized as 0.0025 and 0.0005. The number of constructed local geometric structures is set as 64, and their features are extracted from the third convolutional block in encoder $E$. The optimization objective of our model in the $s$-th incremental state is formally formulated as follows:
\begin{equation}
\min\limits_{\theta_{D_s}}\mathcal{L}_{\mr{obj}}^s = \mathbb{E}_{(x_i^s, y_i^s)\in D_s}[- \sum\limits_{t=1}^{c_p+c_s}(y_i^s)^t\mr{log}(C(f_c; \theta_{C})^t)],
\label{eq:overall_training_objective}
\end{equation}
where $C(f_c; \theta_{C})^t$ and $(y_i^s)^t$ denote the probability predicted as the $t$-th class in the training phase and its corresponding one-hot label, respectively. $\theta_{D_s}$ represents all network parameters of our I3DOL model for simplification, which is composed of $\theta_{E}, \theta_{C}, \theta_{T_p}, \theta_{T_g}, \theta_{T_u}$ and $\theta_{T_d}$. 
Moreover, \textbf{Algorithm 1} summarizes the whole optimization framework of our proposed I3DOL model.

\begin{table*}[t]
	\centering
	\setlength{\tabcolsep}{3.17mm}
	\caption{Quantitative comparisons on ShapeNet dataset \cite{DBLP:journals/corr/ChangFGHHLSSSSX15} with an increment of 6 classes.}
	\scalebox{0.94}{
		\begin{tabular}{|c|ccccccccc|c|}
			\hline
			Comparison Methods & 6 & 12 & 18 & 24 & 30 & 36 & 42 & 48 & 53 & Avg \\
			\hline
			LwF \cite{10.1007/978-3-319-46493-0_37}	& 96.3 & 86.8 & 78.5 & 68.3 & 60.7 & 52.4 & 45.1 & 42.6 & 39.5 & 63.4 \\
			iCaRL \cite{Rebuffi_2017_CVPR} & 96.7 & 88.4 & 82.1 & 74.9 & 68.5 & 62.3 & 56.9 & 51.3 & 44.6 & 69.5 \\
			DeeSIL \cite{deesil-eccv2018} & 97.1 & 90.2 & 84.3 & 76.5 & 73.7 & 65.6 & 57.3 & 53.6 & 47.2 & 71.7 \\
			EEIL \cite{Castro_2018_ECCV} & 97.3 & 91.8 & 86.4 & 79.5 & 73.1 & 67.3 & 63.4 & 57.1 & 51.6 & 74.2  \\
			IL2M \cite{Belouadah_2019_ICCV} & 97.5 & 91.4 & 86.7 & 79.8 & 75.6 & 71.8 & 69.1 & 64.8 & 61.4 & 77.6 \\
			DGMw \cite{Oleksiy_2019_CVPR} & 97.2 & 90.8 & 85.9 & 78.3 & 74.4 & 69.5 & 62.4 & 56.3 & 49.2 & 73.8 \\
			DGMa \cite{Oleksiy_2019_CVPR} & 97.2 & 91.6 & 85.1 & 77.9 & 73.2 & 68.5 & 62.8 & 55.4 & 48.7 & 73.4 \\
			BiC \cite{Wu_2019_CVPR} & 97.4 & 92.1 & 86.7 & 81.5 & 76.4 & 73.7 & 69.8 & 67.6 & 64.2 & 78.8 \\
			RPS-Net \cite{NIPS2019_9429} & \textbf{97.6} & 92.5 & 87.4 & 80.1 & 77.4 & 72.3 & 68.4 & 66.5 & 63.5 & 78.4 \\
			\hline
			\hline
			\rowcolor{lightgray}
			Ours-w/oAG & 97.3 & 92.1 & 87.5 & 80.4 & 76.6 & 72.7 & 69.8 & 65.2 & 62.3 & 78.2 \\
			
			\rowcolor{lightgray}
			Ours-w/oGA & 97.4 & 92.8 & 88.1 & 82.0 & 77.3 & 74.8 & 71.4 & 68.7 & 65.4 & 79.8 \\
			
			\rowcolor{lightgray}
			Ours-w/oSF & 96.6 & 93.8 & 89.1 & 82.8 & 78.0 & 75.6 & 72.8 & 69.5 & 66.4 & 80.5 \\
			
			\rowcolor{lightgray}
			Ours & 97.5 & \textbf{94.4} & \textbf{90.2} & \textbf{84.3} & \textbf{80.5} & \textbf{76.1} & \textbf{73.5} & \textbf{70.8} & \textbf{67.3} & \textbf{81.6} \\
			\hline					
		\end{tabular}
	} 	
	\label{tab:exp_shapenet54_dataset}
\end{table*}

\begin{figure}[t]
	\centering
	\includegraphics[height=160pt, width=235pt]{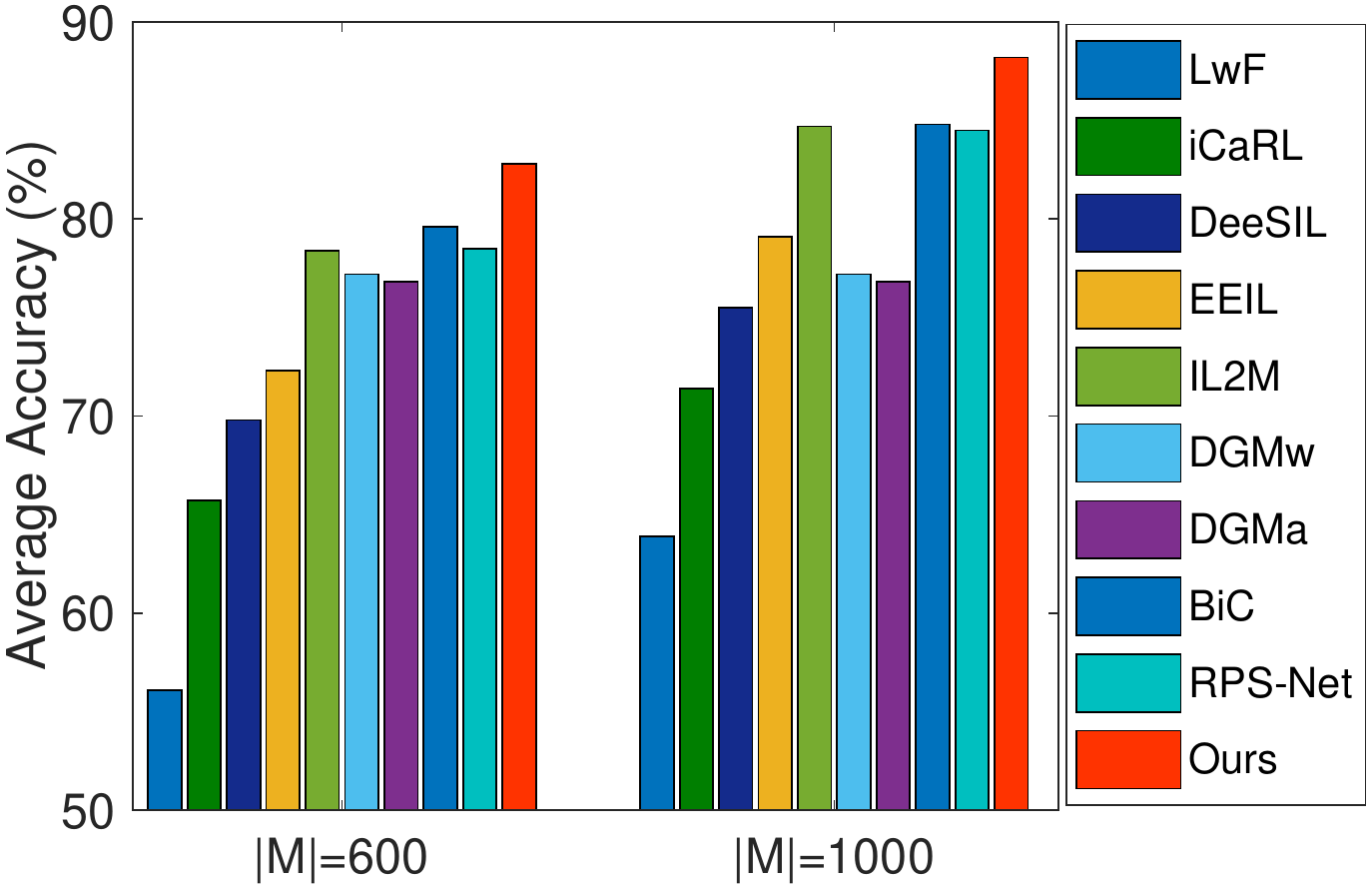}
	\caption{Effect investigation about different number of exemplars on ModelNet dataset.}
	\label{fig:effect_different_exemplars}
\end{figure}

\begin{figure}[t]
	\centering
	\includegraphics[height=200pt, width=235pt]{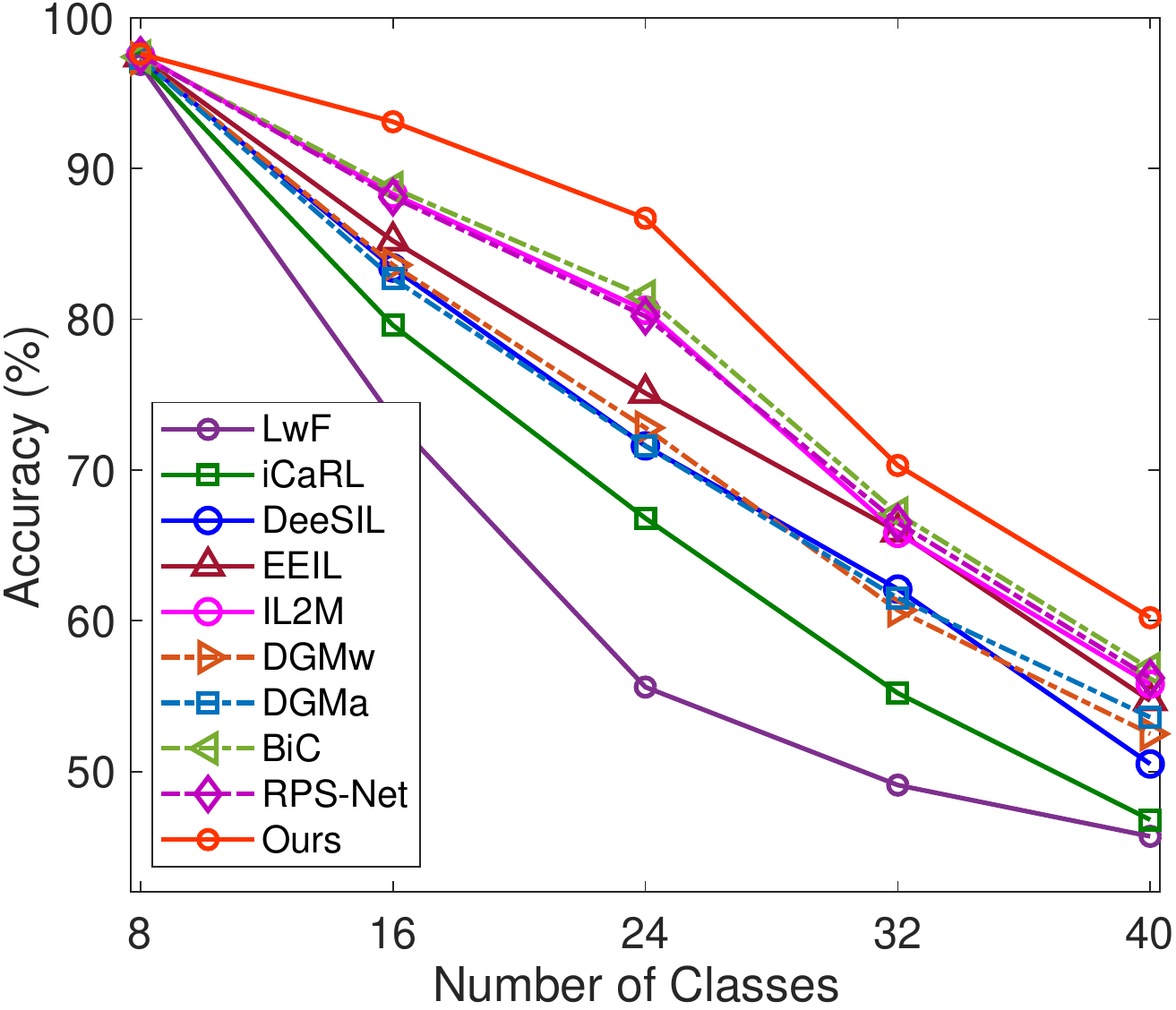}
	\caption{Effect investigation about different number of total incremental states on ModelNet dataset.}
	\label{fig:effect_different_incremental_states}
\end{figure}

\begin{figure}[t]
	\centering
	\includegraphics[height=205pt, width=235pt]{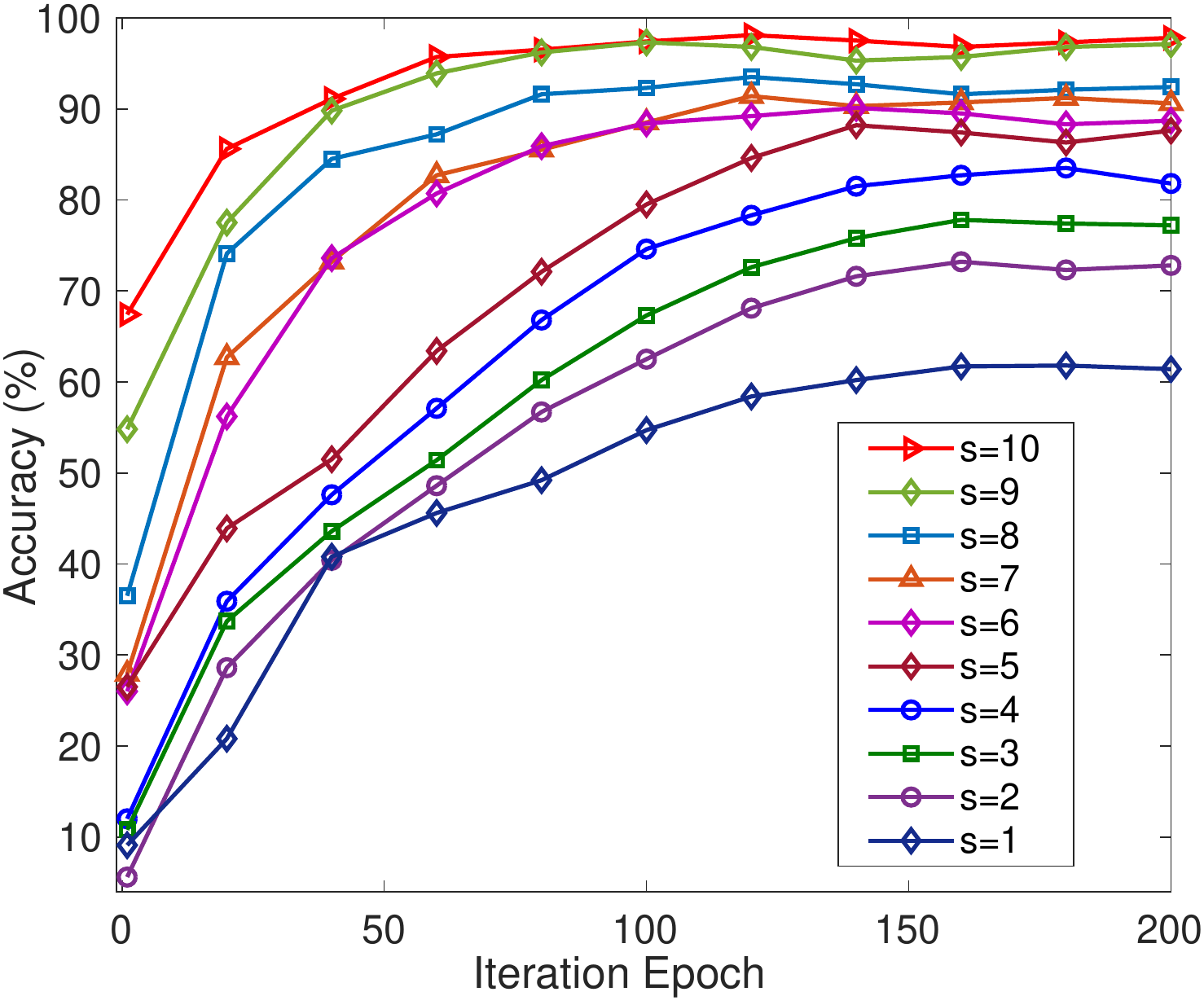}
	\caption{Convergence investigation about all incremental states on ModelNet dataset when $S=10$.}
	\label{fig:convergence_investigation_ModelNet}
\end{figure}

\section{Experiments}
In this section, all advanced comparison approaches utilize PointNet \cite{Qi_2017_CVPR} as baseline feature extractor, and also perform data augmentation for point cloud in the training phase.

\subsection{Datasets and Evaluation}
Generally, three representative point cloud datasets, \emph{i.e.}, ModelNet \cite{7298801}, ShapeNet \cite{DBLP:journals/corr/ChangFGHHLSSSSX15} and ScanNet \cite{8099744} are employed to validate the superiority of our I3DOL model. {ModelNet} \cite{DBLP:journals/corr/ChangFGHHLSSSSX15} consists of 9843 training samples and 2468 testing samples, which are clean 3D CAD models from 40 classes. We select 800 samples as the exemplars set $M$, and set the total incremental states $S$ as 10. ShapeNet \cite{DBLP:journals/corr/ChangFGHHLSSSSX15} contains 35037 training examples and 5053 validation examples. In our experiments, we utilize 53 categories of 3D CAD models that are gathered from online repositories. 1000 samples are stored as the exemplars $M$ and $S$ is set as 9. {ScanNet} \cite{8099744} with 17 categories is composed of scanned and reconstructed real-world indoor scenes, where the training and validation samples are 12060 and 3416, respectively. We set $M=600$ and $S=9$. For performance evaluation, top-1 classification accuracy is employed as basic metric.

\begin{table*}[t]
	\centering
	\setlength{\tabcolsep}{3.17mm}
	\caption{Quantitative comparisons on ScanNet dataset \cite{8099744} with an increment of 2 classes.}
	\scalebox{0.94}{
		\begin{tabular}{|c|ccccccccc|c|}
			\hline
			Comparison Methods & 2 & 4 & 6 & 8 & 10 & 12 & 14 & 16 & 17 & Avg \\
			\hline
			LwF \cite{10.1007/978-3-319-46493-0_37}	& 92.2 & 74.8 & 60.3 & 48.2 & 41.6 & 37.3 & 35.7 & 33.5 & 31.8 & 53.1  \\
			iCaRL \cite{Rebuffi_2017_CVPR} & 92.4 & 78.7 & 67.4 & 59.7 & 52.5 & 48.2 & 43.5 & 39.9 & 36.3 & 56.0 \\
			DeeSIL \cite{deesil-eccv2018} & 92.6 & 80.1 & 71.5 & 63.3 & 57.3 & 52.8 & 48.6 & 45.2 & 43.7 & 63.1 \\
			EEIL \cite{Castro_2018_ECCV} & 92.7 & 83.4 & 75.6 & 72.6 & 58.7 & 55.4 & 52.3 & 49.4 & 45.7 & 65.1 \\
			IL2M \cite{Belouadah_2019_ICCV} & 92.9 & 84.4 & 77.3 & 70.1 & 60.8 & 56.7 & 54.1 & 52.6 & 48.3 & 66.7 \\
			DGMw \cite{Oleksiy_2019_CVPR} & 92.5 & 82.6 & 67.1 & 61.8 & 56.3 & 53.2 & 50.8 & 47.5 & 43.8 & 63.6  \\
			DGMa \cite{Oleksiy_2019_CVPR} & 92.5 & 82.2 & 67.8 & 60.2 & 56.6 & 52.7 & 50.4 & 48.1 & 44.7 & 63.7  \\
			BiC \cite{Wu_2019_CVPR} & 92.8 & 84.2 & 77.5 & 70.3 & 60.6 & 57.2 & 54.3 & 52.4 & 48.5 & 66.8 \\
			RPS-Net \cite{NIPS2019_9429} & 92.9 & 84.8 & 77.1 & 70.7 & 61.2 & 57.6 & 55.4 & 53.3 & 49.1 & 67.3  \\
			\hline
			\hline
			\rowcolor{lightgray}
			Ours-w/oAG & 92.9 & 83.7 & 76.8 & 73.6 & 60.2 & 56.7 & 54.0 & 51.8 & 47.5 & 66.4  \\
			
			\rowcolor{lightgray}
			Ours-w/oGA & 93.2 & 85.5 & 77.9 & 75.6 & 62.1 & 58.3 & 56.4 & 53.9 & 51.0 & 68.2   \\
			
			\rowcolor{lightgray}
			Ours-w/oSF & \textbf{93.3} & 86.1 & 78.5 & 76.0 & 62.6 & 59.1 & 56.3 & 54.8 & 51.5 & 68.7 \\
			
			\rowcolor{lightgray}
			Ours & 93.2 & \textbf{87.2} & \textbf{80.5} & \textbf{77.8} & \textbf{64.3} & \textbf{61.9} & \textbf{58.2} & \textbf{56.8} & \textbf{52.1} & \textbf{70.2} \\
			\hline					
		\end{tabular}
	} 	
	\label{tab:exp_scannet17_dataset}
\end{table*}

\subsection{Experiments on ModelNet Dataset}
\subsubsection{Performance Comparisons}
We present the comparison performance on ModelNet dataset \cite{7298801} in Table~\ref{tab:exp_modelnet40_dataset}, with an increment of 4 classes for each incremental state. Some key observations from the results in Table~\ref{tab:exp_modelnet40_dataset} can be summarized as follows: 1) Our proposed I3DOL model significantly outperforms all advanced comparison approaches \cite{Rebuffi_2017_CVPR, Castro_2018_ECCV, Belouadah_2019_ICCV, Oleksiy_2019_CVPR, Wu_2019_CVPR, 10.1007/978-3-319-46493-0_37, NIPS2019_9429} in 2D vision about 3.6\%$\sim$25.2\% in terms of average accuracy, which illustrates the superiority of our I3DOL model. 2) For classes incremental learning of 3D objects, our model effectively alleviates the catastrophic forgetting for past classes of 3D objects when comparing with other competing approaches \cite{Rebuffi_2017_CVPR, Castro_2018_ECCV, Belouadah_2019_ICCV, Oleksiy_2019_CVPR, Wu_2019_CVPR, NIPS2019_9429}. 3) The irregular point cloud representation can be characterized well via our I3DOL model to promote the classification performance.

\subsubsection{Ablation Studies}
As the gray part presented in Table~\ref{tab:exp_modelnet40_dataset}, empirical variant experiments on ModelNet dataset are prepared to illustrate the effectiveness of different components in our I3DOL model. Moreover, we respectively denote our proposed I3DOL model without only adaptive-geometric centroid construction, geometric-aware attention mechanism and score fairness compensation as Ours-w/oAG, Ours-w/oGA and Ours-w/oSF. The average prediction accuracy degrades 1.2\%$\sim$3.8\% when any component is abandoned from our proposed I3DOL model. Furthermore, it demonstrates that all proposed modules play an indispensable role in highlighting unique and informative 3D geometric characteristics with high contributions for classes incremental learning of 3D objects.

\subsubsection{Effects of Exemplars and Incremental States}
As shown in Figure~\ref{fig:effect_different_exemplars} and Figure~\ref{fig:effect_different_incremental_states}, this subsection investigates the effects of different exemplars and incremental states on ModelNet dataset by setting different values of $|M|$ and $S$. Specifically, some essential conclusions drawn from the results in Figure~\ref{fig:effect_different_exemplars} and Figure~\ref{fig:effect_different_incremental_states} are summarized as follows: 1) Our proposed I3DOL model could better prevent the catastrophic forgetting for past classes of 3D objects in different settings of exemplars and incremental states. 2) More selected exemplars encourage our I3DOL model to better alleviate the catastrophic forgetting for past learned 3D classes brought by redundant 3D geometric characteristics and unbalanced training samples.

\subsubsection{Convergence Analysis}
Figure~\ref{fig:convergence_investigation_ModelNet} investigates the convergence stability of our model on ModelNet dataset. Specifically, our proposed I3DOL model presents the stable performance when the iterative training epoch is about 140. Moreover, our I3DOL model could achieve efficient convergence across all incremental states.

\subsection{Experiments on ShapeNet and ScanNet Datasets}
As shown in Table~\ref{tab:exp_shapenet54_dataset} and Table~\ref{tab:exp_scannet17_dataset}, this subsection introduces extensive quantitative comparisons and ablation studies on ShapeNet \cite{DBLP:journals/corr/ChangFGHHLSSSSX15} and ScanNet \cite{8099744}. Some conclusions can be drawn from the presented comparison performance: 1) When compared with other advanced comparison approaches such as \cite{Rebuffi_2017_CVPR, Castro_2018_ECCV, Wu_2019_CVPR, NIPS2019_9429, 10.1007/978-3-319-46493-0_37}, our I3DOL model achieves better performance to alleviate catastrophic forgetting, which improves 2.8\%$\sim$18.2\% in terms of average accuracy. 2) Ablation studies verify that each component of our I3DOL model is designed effectively to facilitate classes incremental learning of 3D objects. 3) Our I3DOL model could better explore unique and informative 3D geometric characteristic, and address the unbalanced data distributions to alleviate catastrophic forgetting for past learned classes of 3D objects.

\section{Conclusion}
In this paper, we develop a new Incremental 3D Object Learning (\emph{i.e.}, I3DOL) model to continually explore new classes of 3D objects via alleviating catastrophic forgetting for past classes. Specifically, the adaptive-geometric centroid module is used to construct several discriminative local geometric structures to characterize the irregular point cloud representation. Meanwhile, the geometric-aware attention mechanism highlights unique and informative 3D geometric characteristics with high contributions for classes incremental learning of 3D objects. Moreover, we propose the score fairness compensation strategy to correct biased score prediction, which effectively prevents the catastrophic forgetting of past classes. The effectiveness of our I3DOL model is justified well via extensive experiments.

\bibliography{I3DOL-1924}

\end{document}